\pdfoutput=1

\documentclass[11pt]{article}

\usepackage{EMNLP2023}

\usepackage{times}
\usepackage{latexsym}

\usepackage[T1]{fontenc}

\usepackage[utf8]{inputenc}

\usepackage{microtype}

\usepackage{inconsolata}

\usepackage{graphicx}

\title{How Good Are SOTA Fake News Detectors?}

\author{Matthew Iceland \\
  University of Rochester \\
  \texttt{miceland@u.rochester.edu} \\
}

\begin{document}
\maketitle
\begin{abstract}
Automatic fake news detection with machine learning can prevent the dissemination of false statements before they gain many views. Several datasets labeling news statements as legitimate or false have been created since the 2016 United States presidential election for the prospect of training machine learning models. We evaluate the robustness of both traditional and deep state-of-the-art models to gauge how well they may perform in the real world. We find that traditional models tend to generalize better to data outside the distribution it was trained on compared to more recently-developed large language models, though the best model to use may depend on the specific task at hand.
\end{abstract}

\section{Introduction}

\citet{Zhou_2020} broadly defined fake news as news that is false. While the intentions and motivations behind fake news are less well-defined, a key characteristic is that it can be disproven through fact-checking. However, many associate the term, "fake news," with intentional efforts to spread lies. Fake news is a threat to democracy and fair elections, and it is capable of causing significant disruptions in financial markets \citep{Fong2021}. Automatic detection of false news using state of the art (SOTA) natural language processing techniques would help mitigate its spread and harmful impacts.

While machine learning models are fallible and may label a false story as true or vice-versa, they can automatically filter out many of the stories that humans must manually fact-check. Reducing the number of statements that must be fact-checked is especially important today as social media platforms enable fake news to be generated and spread very easily.

The goal of this study is to evaluate how well different machine learning models may perform when deployed in the real world. Crucial to strong real-word performance is the capability of fake news detectors to generalize well to data outside the distribution it was trained on. We hope to learn if SOTA models are capable of learning patterns that apply to text outside the dataset it was trained on, or if they just learn nuances specific to the training dataset. We only train machine learning models on article titles or single news statements, yet such information alone may be sufficient to classify fake news with high accuracy and F1 scores.

We would also like to clearly distinguish between how traditional machine learning algorithms, such as Naive Bayes and random forest, perform in comparison to deep language transformers, such as BERT \citep{devlin2019bert}. Traditional models have the benefit of being much less computationally expensive and offer more explainability in how they make decisions. Hence, we would like to explore whether the increased complexity and training cost of transformers yields benefits other than higher benchmark performances on exiting datasets.

Our three main research questions are as follows:
\begin{enumerate}
    \item How well do fake news detectors perform on out-of-sample datasets?
    \item How accurately can fake news detectors identify neurally generated fake news with similar content but different style from which they were trained on?
    \item How do traditional machine learning models perform compared to deep learning models?
\end{enumerate}

Across all of our experiments, deep learning models tend to achieve higher accuracy and F1 scores when classifying fake news within the same distribution it was trained on. However, we find that traditional machine learning models generalize better to out-of-sample datasets and AI-synthesized fake news, though no single model clearly outperforms the others across all of our tests.

\section{Related Works}

Existing studies \citep{info13120576, Khan_2021, inproceedings, gundapu2021transformer} indicate that contextual embeddings produced by transformers such as BERT tend to result in higher classification accuracies for fake news datasets than traditional or deep models combined with non-contextual embeddings, such as Word2Vec and GloVe. For example, \citet{info13120576} reports that a BERT\textsubscript{BASE} with a fully-connected classifier achieves an accuracy of 91.37\% on the FakeNewsNet \citep{shu2018fakenewsnet, shu2017fake, shu2017exploiting} dataset, while the best model using GloVe or Word2Vec embeddings achieved only 82.01\% accuracy. Also, \citet{Khan_2021} reports that a BERT\textsubscript{BASE} and a RoBERTa\textsubscript{BASE} with a fully-connected classifier each scored an accuracy of 62\% on the LIAR \citep{wang-2017-liar} dataset for binary classification (i.e. where the "pants-fire", "false", and "barely-true" statements are combined into a single fake category and the "half-true", "mostly-true", and "true" are combined into a single real category) while the best non-transformer model was Naive Bayes with TF-IDF embeddings, reaching an accuracy of 60\%.

\citet{aich-etal-2022-demystifying} compares the importance of different stylistic features (e.g. punctuations, negations, misspelled words), complexity features (e.g. word count and word length), and psychological features (e.g. polarization), in classifying fake news. They create a collection of fake news articles using a generative adversarial network and combine them with 30 real news articles from Buzzfeed. They extract 21 different stylistic, complexity, and psychological features from these articles to train a K nearest neighbor classifier. They find that stylistic features, and specifically the number of misspelled words, out-of-vocabulary words, stop-words, proper nouns, and camel-case words, are most useful for accurate fake news classification.

\citet{Hays_2023} evaluates robustness and versatility of state-of-the-art models for the similar task of bot detection. They suggest that SOTA methods for discriminating between bots and humans achieve high accuracies due to limitations in the datasets used to evaluate them. They find that shallow decision trees can often achieve roughly the same accuracy as SOTA methods for existing datasets, and some widely-used datasets can be classified by asking simple yes/no questions, such as whether an account has ever tweeted the word, "earthquake," or liked more than 16 tweets. Furthermore, when training a random forest classifier on each of their studied datasets, it only achieved about 50\% accuracy on most out-of-sample datasets. The results of \citet{Hays_2023} also raises concerns over the quality of fake news detection datasets, for some are often classified with near-perfect accuracy in past studies. For example, \citet{SAMADI2021102723} achieves accuracies of over 99\% on the ISOT Fake News Dataset \citep{inproceedingsAhmed} using BERT, RoBERTa \citep{liu2019roberta}, GTP2 \citep{noauthororeditor}, and Funnel-transformers \citep{dai2020funneltransformer} with linear and CNN classifiers.

\section{Experiment Setup}

The datasets and models used in our study are decribed below. We use five different publicly-available fake news datasets, and we consider six classical machine models and six commonly-used language transformers. All classical models are implemented using the Python \texttt{sklearn} library except for the gradient boosting algorithm, which is implemented with the \texttt{xgboost} library. TF-IDF word embeddings are used because they often work well with traditional algorithms for fake news detection \citep{9620068}. All transformers are implemented with the \texttt{transformers} library, and the classification version of each model is used (e.g \texttt{BertForSequenceClassifcation}), which adds a linear classifier on top of the output of the \texttt
{CLS} token. All experiments were performed on a system with 64GB of RAM, an Intel Core i9 CPU, and an NVIDIA GeForce RTX 4090 GPU. Our code is available at https://github.com/miceland2/Fake\_news\_detection.

\subsection{Datasets}

\begin{table}
  \centering
  \caption{The number of true and fake articles in each dataset and their percent train and test splits}
  \begin{tabular}{|l|l|l|l|}
    \hline
    Dataset & True & Fake & Train / Test \\
    \hline
    D1 & 21,417 & 23,481 & 80 / 20 \\
    \hline
    D2 & 4,507 & 3,554 & 75 / 25 \\
    \hline
    D3 & 9,855 & 10,387 & 80 / 20 \\
    \hline
    D4 & 17,441 & 5,755 & 80 / 20 \\
    \hline
    D5 & 2,061 & 1,057 & 80 / 20 \\
    \hline
        
  \end{tabular}
\end{table}

The ISOT Fake News Dataset \citep{inproceedingsAhmed} contains about 45 thousand articles, which include both titles and body text. Most of the articles are about government and politics. The truthful news articles are collected from Reuters.com, and the dishonest news articles come from a variety of websites that were flagged by PolitiFact and WikiPedia as unreliable.

The LIAR dataset \citep{wang-2017-liar} contains 12.8 thousand short statements that were collected and manually labeled over a decade; the dataset was released in 2017. The statements come from PolitiFact and consist of roughly an equal amount of Democrat and Republican speakers. The dataset classifies the statements into six categories: pants-fire, false, barely true, half-true, mostly true, and true. LIAR is known to be a rather difficult dataset, for its original paper \citep{wang-2017-liar} achieves only 27.4\% accuracy using a hybrid CNN trained on the statements combined with other contextual information, such as the subject and speaker. Because we are interested in evaluating models on out-of-sample data, we implement a binary version of the original LIAR dataset, where the "true" and "mostly true" labels are combined into a single truthful category, and the "pants-fire" and "false" labels are combined into a single untruthful category. Entries with the "barely true" or "half true" label are dropped.

Our third dataset \citep{fake-news} comes from kaggle and was published in 2018. Entitled "Fake News," it contains about 20 thousand entries which are each labeled as reliable or unreliable. The dataset comes with both a title and body text for each article.

The FakeNewsNet dataset \citep{shu2018fakenewsnet, shu2017fake, shu2017exploiting} consists of political articles labeled by PolitiFact and entertainment articles evaluated by GossipCop. However, about 22 thousand of the articles in the repository come from GossipCop while only about a thousand come from Politifact. Like the LIAR dataset, the FakeNewsNet entries don't come with full body texts for each statement, but instead a single title that is real or fake.

Our final dataset \citep{covid-fake-news} contains true and fake news articles related to the topic of COVID-19. The data was collected from December 2019 to July 2020 and manually labeled into three categories: false, partially false, and true. For our binary classification task, the "partially false" and "false" entries are combined into a single category.

In Table 1, we summarize the datasets with their number of true and fake examples as well as the ratio between the sizes of their training and test sets. For ease of labeling, we refer to the ISOT Fake News dataset as D1, LIAR as D2, the kaggle "Fake News" dataset as D3, FakeNewsNet as D4, and the COVID-19 Fake News dataset as D5. We observe that the first three datasets are roughly balanced in terms of their real/fake news ratio, but D4 and D5 are made up of about 75\% and 66\% true news, respectively.

\subsection{Deep Learning Models}

We utilize six commonly-used transformers for binary sequence classification. We use the "Base" size for each model below unless otherwise specified, and they are fine-tuned for ten epochs on each of the datasets.

BERT \citep{devlin2019bert}, which stands for Bidirectional Encoder Representations from Transformers, was one of the first great successes for the transformer \citep{vaswani2017attention} model in natural language processing. One of the defining mechanisms of the transformer is self-attention, which allows the model to more easily learn long-range dependencies compared to recurrent or convolutional neural networks \citep{vaswani2017attention}. Natural language transformers like BERT find the relations between all tokens in the input sequence to generate context-aware word embeddings. For a BERT\textsubscript{BASE}, each of the input tokens is represented as a 768-dimensional vector, computed using 12 layers of 12 self-attention heads each. BERT models are pre-trained in an unsupervised manner using the next sentence prediction (NSP) objective and the masked language model (MLM) objective, which masks some of the input tokens and forces the model to recover their vocabulary id using only their context. BERT models are efficient enough to be fine-tuned on a single GPU.

RoBERTa \citep{liu2019roberta} and DebBERTa \citep{he2021deberta} are improved variations of BERT that that also use the MLM pre-training objective and outperform the original model on natural language processing tasks such as GLUE, RACE, and SQuAD. COVID-Twitter-BERT \citep{müller2020covidtwitterbert}, also known as CT-BERT, utilizes the BERT\textsubscript{LARGE} architecture but is pretrained on a corpus of 22.5 million tweets specifically about the coronavirus. The maximum sequence length is only 96 instead of 512 for the original BERT model, though this length suffices for virtually all statements in the datasets except four from LIAR, which are excluded from our experiments.

ELECTRA \citep{clark2020electra} is another model based off the BERT architecture that uses a pre-training task called replacement token detection instead of the masked language model. Under this objective, the model learns to discriminate between original and generated tokens in a corrupted sequence. ELECTRA rivals RoBERTa and XLNet in performance but requires much less compute to train.

Finally, XLNet \citep{yang2020xlnet} is based on a separate transformer architecture from BERT that learns bidirectional contexts through an autoregressive pretraining method. This is meant to overcome BERT's MLM objective limitation where it assumes the predicted tokens are idenpendent of each other.

\subsection{Classical Models}

Compared to recent deep learning methods for sequence classification, classical machine learning models are much more simple and inexpensive to train. In order to convert each statement into a numerical form, we use TF-IDF word embeddings, which aim to capture the statistical importance of each of the words. We implement TF-IDF embeddings using the \texttt{TfidfVectorizer} class from the \texttt{sklearn} library, and terms that occur less than ten times in each set of documents are ignored when building the vocabulary. This class computes the TF-IDF of a term $t$ in document $d$ using the formula

\[
tf(t, d) \times [\log{\frac{n}{df(t)}} + 1],
\]

where $tf(t, d)$ is the term frequency of term $t$ in document $d$, $df(t)$ is the document frequency of term $t$, and $n$ is the number of different documents. The six different classifiers we use with these embeddings are Naive Bayes, support vector machine, adaptive boosting, gradient boosting, random forest, and K nearest neighbor.

Naive Bayes (NB) is a probabilistic classification algorithm that ("naively") assumes the conditional probabilities of each pair of features, given the class label, are independent. For each statement, the algorithm chooses the class label $y$ that maximizes the expression

\[
P(y) \prod_{i=1}^{n} P(x_i \mid y),
\]

where each $x_i$ is a feature of the statement. We use multinomial Naive Bayes to estimate the probabilities above.

A support vector machine (SVM) aims to find a optimal hyperplane to separate the data into classes. We use an RBF kernel for our classifier.

Boosting is a general machine learning technique that aims to build a strong model from several weak ones. Adaptive boosting (AdaBoost) sequentially generates weak models by increasing the importances of the missed training examples from the previous model, and more accurate models are given greater weight when voting on the final classification. In our case, each weak model is a decision tree with a depth of one, and a maximum of 50 trees are created before training terminates. Gradient boosting is a an algorithm where a weak decision tree is initially created, and new trees are sequentially added to predict the errors of the previous model. XGBoost, which stands for "extreme gradient boosting," is a library that provides an implementation of gradient boosting with regularization to reduce overfitting.

Random forest (RF) is another ensembling method that makes use of voting by several decision trees. Random forest is a bagging algorithm where random samples of the training data are drawn with replacement to train each decision tree. The algorithm makes use of feature randomness to generate a random subset of features for each tree to ensure low correlation between them. We use 100 decision trees with no limit on depth and a minimum of two samples required to split an internal node.

Lastly, K nearest neightbor (k-NN) operates off the assumption that similar points are closee to each other in the feature space. Each data point, or statement in our case, is classified based on the most frequent label of the K training points closest to it. We use the \texttt{sklearn} default of the top five closest neighbors to a given point.

\section{Results}

\subsection{Preliminary}

\begin{table}
  \centering
  \caption{Accuracy score for each model trained and tested on each dataset}
  \begin{tabular}{|l|l|l|l|l|l|}
    \hline
    Model & D1 & D2 & D3 & D4 & D5 \\
    \hline
    \hline
    BERT & 98.7 & 63.0 & 96.0 & \textbf{85.3} & 75.0 \\
    \hline
    RoBERTa & \textbf{99.9} & 67.4 & - & 82.1 & 77.9 \\
    \hline
    DeBERTa & \textbf{99.9} & 67.5 & 96.1 & \textbf{85.3} & 75.5 \\
    \hline
    CT-BERT & \textbf{99.9} & \textbf{69.7} & \textbf{96.4} & 85.0 & \textbf{80.3} \\
    \hline
    ELECTRA & 98.5 & 66.7 & 95.3 & 82.2 & 76.9 \\
    \hline
    XLNet & \textbf{99.9} & 65.9 & 96.2 & 84.4 & 77.4 \\
    \hline
    \hline
    NB & 92.7 & 64.0 & 85.9 & 82.8 & 70.5 \\
    \hline
    RF & 93.2 & 63.7 & 93.5 & 82.8 & 73.7 \\
    \hline
    AdaBoost & 84.9 & 60.3 & 91.9 & 80.6 & 70.2 \\
    \hline
    XGBoost & 90.7 & 62.2 & 92.7 & 82.2 & 72.6 \\
    \hline
    k-NN & 86.9 & 50.9 & 56.2 & 76.9 & 68.9 \\
    \hline
    SVM & 95.2 & 64.4 & 92.9 & 84.0 & 73.2 \\
    \hline
        
  \end{tabular}
\end{table}

We first simply train each of the 12 models on each of the five datasets. Our two measures of performance are accuracy and F1. Accuracy can be simply defined as the number of true positives and true negatives over the total number of testing examples. We also consider F1 score, which is defined as the harmonic mean between precision and recall, or, equivalently, by the expression

\[
\frac{TP}{TP + \frac{1}{2}(FP + FN)}.
\]

Our preliminary results are summarized in Table 2. We report accuracy for each model and include F1 scores in the following section. The language transformers clearly outperform the traditional models across all datasets. All the best models for the five datasets are deep language transformers, and CT-BERT in particular stands out as one of the best performers. We also observe that, unlike the traditional models, the deep models are able to achieve near-perfect performance on D1.

\subsection{Out-of-Sample Data}

Fake news detectors will only be effective in the real world if they're able to generalize to out-of-sample data. As noted by \citet{Hays_2023}, strong performance on a given dataset may not necesarily be the result of model robustness; rather, it may be a consequence of limitations in the data collection process. We take our trained models from Section 4.1 and evaluate them with the four other test sets. For the sake of space, we only display the $5 \times 5$ matrices for the two best deep learning models and the two best traditional models in terms of overall generalization performance, and we include the remaining matrices in Appendix A. Each row signifies the dataset the model was trained on, and the column signifies the dataset used for evaluation. We use the F1 score of each model to evaluate the performance in each of the train-test cases.

The matrices for ELECTRA and XLNet are shown in Figure 1. The entries along the main diagonal tend to be the darkest, representing the cases when the models are trained and tested on the same dataset. However, the F1 scores of most cells off the main diagonal are close to 50, which is the equivalent of random guessing. The main exceptions are the third rows when the models are trained on the kaggle "Fake News" dataset (D3), plus a few other combinations of train-test pairs. Even when one of the transformers does better than random guessing on a test set it wasn't trained on, the drops in F1 scores are usually substantial.

\begin{figure}
    \includegraphics[width=.43\textwidth]{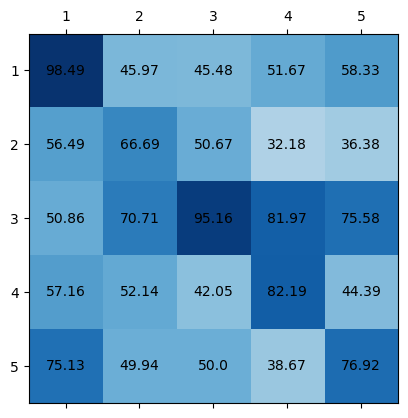}\hfill
    \includegraphics[width=.43\textwidth]{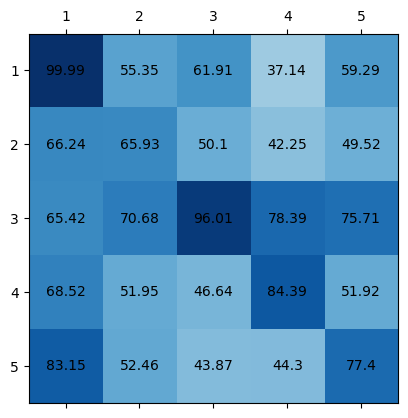}\hfill
    \caption{Out-of-sample performance matrices for ELECTRA (top) and XLNet (bottom)}\label{fig:foobar}
\end{figure}

\begin{figure}
    \includegraphics[width=.43\textwidth]{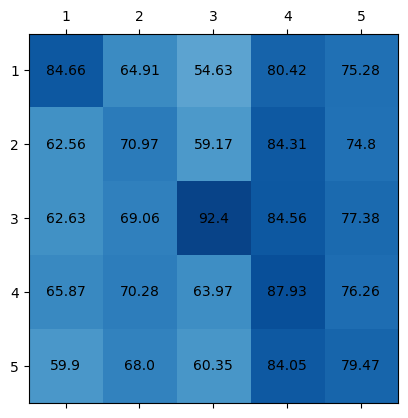}\hfill
    \includegraphics[width=.43\textwidth]{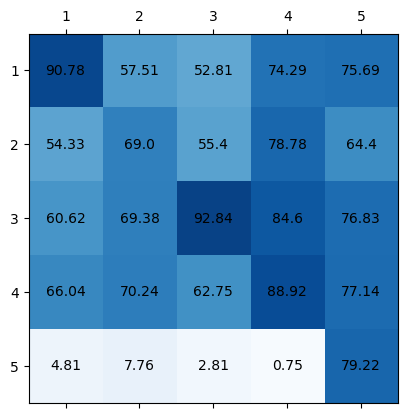}\hfill
    \caption{Out-of-sample performance matrices for AdaBoost (top) and XGBoost (bottom)}\label{fig:foobar}
\end{figure}

\begin{table}
  \centering
  \caption{Percent of titles in the validation sets for D4 and D5 that include feature used in AdaBoost trained on a separate dataset}
  \begin{tabular}{|l|l|l|}
    \hline
    Dataset & \% of D4 titles & \% of D5 titles \\
    \hline
    D1 & 20.35 & 25.80 \\
    \hline
    D2 & 14.12 & 12.50 \\
    \hline
    D3 & 15.39 & 12.34 \\
    \hline
        
  \end{tabular}
\end{table}

The two boosting algorithms, AdaBoost and XGBoost, are overall the best performers on the out-of-sample task. The matrices in Figure 2 are clearly more homogeneous in shade, reflecting their less severe drops in performance when trained on out-of-distribution data. The only exception is when XGBoost is trained on the COVID-19 dataset (D5), in which case the model generalizes extremely poorly. However, this dataset may be designed for classification specific to the coronavirus domain. The cells of all other rows are at least above 50, though a majority are reasonably better than random guessing. Some models even perform better on the out-of-sample test sets, such as both boosting algorithms trained on LIAR (D2) and evaluated on FakeNewsNet (D4).

The matrices of the other four transformers look similar to those of ELECTRA and XLNet but generally have even lower F1 scores off the main diagonal. Random forest performs poorly overall on out-of-sample data while the matrices of the other traditional models are more homogenous in shade.

\subsection{Explaining the AdaBoost Performance}

We take a closer look into how the AdaBoost classifiers make their decisions, for they they tend to generalize best to unseen datasets. Each model is composed of 50 decision trees with only a single branch, and gini loss is used to train each of them. We note from Figure 2 that the classifiers trained on D1, D2, and D3 perform especially well when tested on D4 and D5. Thus, we analyze each of the 50 words used by the classifiers of the first three datasets and report the percentage of titles in D4 and D5 that contain at least of one these words. For D1, D2, and D3, we report these percentages in Table 3.

We also include a complete list of the words used by each classifier and the respective training error of each classifier in Appendix B. From Table 3, we can see that only small percentages of the titles of D4 and D5 contains at least one of the words used as a feature in the AdaBoost classifiers. The greatest overlap is between D1 and D5, where 25.8\% of the statements in the latter have words used by the D1 classifier. However, all F1 scores of the AdaBoost classifiers evaluated on D4 and D5 (i.e. the rightmost two columns of top matrix of Figure 2) are at least 74.8. Recall that D4 and D5 are imbalanced, with more real news than fake news, so the AdaBoost models evaluated on these datasets must be predicting true for many statements. Therefore, these high performances suggest that the presence of many of the feature words is indicative of fake news.

Several of these words, listed in Appendix B, can be considered "loaded" terms, such as names of famous politicians or race-related words, though most of the decision tree classifiers have accuracies just barely above 50\%. Many of the words, however, are neutral and less obvious indicators of fake news. For example, the TF-IDF of the word, "video," is used by the ensembles for both D1 and D3 and is used by the most accuracte decision tree for D1.

\subsection{AI-Generated Fake News}

\begin{figure}
    \includegraphics[width=.45\textwidth]{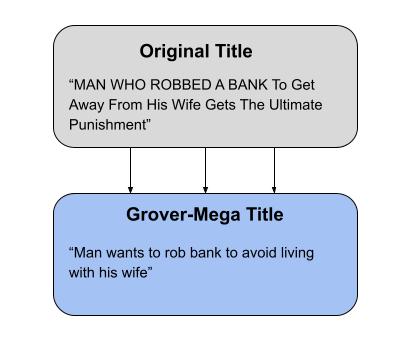}\hfill
    \caption{Example title generated by Grover-Mega from the ISOT Fake News Dataset}\label{fig:foobar}
\end{figure}

For the generation of neural fake use, we utilize Grover \citep{zellers2020defending}, a generative model that can, for our purposes, generate a fake news title given the body text, authors, domain, and date of publication of an article. Grover uses the same architecture as GPT2 \citep{noauthororeditor}; we specifically use the Grover-Mega version, which has 48 layers and about 1.5 billion parameters. For datasets that provide both a fake news title and body text for each entry, we hope to generate a new set of titles from the corresponding body text. These new AI-synthesized titles will ideally have a different style but similar subject matter as the original titles.

For D1 and D5, we use Grover-Mega to generate these new titles. For each title in these datasets, we pass into Grover-Mega the corresponding body text, and we select a domain uniformly at randomly to be "techcrunch.com", "vancouversun.com", "chicagotribune.com", "esquire.com", "mashable.com", "theguardian.com", "theatlantic.com", or "slate.com". The publication date for all generations is arbitrarily chosen to be April 19th, 2019. All the original and generated titles are available on our Github\footnote{\url{https://github.com/miceland2/Fake_news_detection}} page.

As mentioned in Section 3.1, D3 also comes with body text for each article, though we don't include the dataset in this section because many of the body texts are encoded to hundreds or thousands of tokens beyond the limit that Grover was trained on. We continue to use the same training sets as Section 4.1 to train our models, and each Grover-generated title in the new test sets was produced using the body texts of the titles in the original validation sets. A sample title generated by Grover-Mega is shown in Figure 3.

\begin{table}
  \centering
  \caption{Accuracy scores for classifying fake titles generated by Grover-Mega}
  \begin{tabular}{|l|l|l|}
    \hline
    Model & D1 & D5 \\
    \hline
    \hline
    BERT & 59.60 & 56.65 \\
    \hline
    RoBERTa & 50.38 & 49.79 \\
    \hline
    DeBERTa & 49.82 & 35.62 \\
    \hline
    CT-BERT & 51.86 & 69.53 \\
    \hline
    ELECTRA & 56.74 & 55.36 \\
    \hline
    XLNet & 46.62 & 46.78 \\
    \hline
    \hline
    NB & 82.91 & 14.16 \\
    \hline
    RF & 73.07 & 82.40 \\
    \hline
    AdaBoost & 46.64 & 16.74 \\
    \hline
    XGBoost & 48.59 & \textbf{86.70} \\
    \hline
    KNN & \textbf{88.72} & 32.62 \\
    \hline
    SVM & 74.32 & 39.48 \\
    \hline
  \end{tabular}
\end{table}

The results for D1 and D5 are presented in Table 3. Because our new test sets only consist of fake examples, we report the accuracies of the models because the F1 scores would all be zero. For both datasets, almost all of the transformer models perform about equivalent to random guessing. The only exception is CT-BERT trained on the D5, the COVID-19 dataset, which achieves an accuracy of 69.53\%. Overall, however, the traditional models perform much better than the deep models. For example, the K nearest neighbor classifier reaches an accuracy of 88.72\% on D1 while XGBoost reaches an accuracy of 86.70\% on D5. Also, the random forest classifier should be distinguished for achieving reasonable accuracy on both datasets, unlike any of the other models.

\section{Conclusion}

While large language models can achieve the greatest performance when evaluated within the same distribution of existing datasets, we argue that they may not necessarily be more robust than traditional models if deployed in the real world. High or near-perfect performance on fake news datasets may paint a misleading picture of how good SOTA fake news detectors are. In reality, transformer-based fake news detectors seem to be very sensitive to shifts in the data distribution they were trained on. Even evaluating a deep language model on out-of-sample data in the same domain it was trained on, such as politics, can result in performance that is about equivalent to random guessing.

Traditional models are better at capturing general patterns in fake news across different datasets and offer greater clarity as to how they are learned. Nevertheless, traditional models may still suffer substantial decreases in performance when train on out-of-sample data, sometimes being relegated to mere weak classifiers that are only slightly better than random guessing. Also, there is no single traditional algorithm that clearly outperforms the others when tested on data that is AI-generated or comes from foreign datasets. For example, AdaBoost generalizes the best across the five existing datasets but fails for both sets of fake news produced by Grover.

In order to classify fake news reliably, an ensemble of several traditional machine learning classifiers may achieve strong performance. Traditional models would also be easier to ensemble compared to large language models because of their fast performance and low training cost. Alternatively, continual learning could be used as a strong fake news detection framework in which models can incrementally acquire knowledge over long periods of time and adapt to changing data distributions.

\section*{Limitations}


Our paper only considers the classification of single news statements or titles. We do not incorporate the full text of existing articles into our classification largely due to the difficulty in scaling to long texts. For example, transformer encoders like BERT can only process sequences of up to 512 tokens, and many of the original words may have to be broken up into multiple tokens due to vocabulary limitations. Furthermore, we do not consider the more advanced task of multimodal classification with text and images.

Limitations in our computing resources, coupled with the recent adoption of very large language models in industry, also limited the number of SOTA fake news detectors we could evaluate. We limit our study to transformers that can be fine-tuned for text classification in a reasonable amount of time on a single GPU. All our studied models are under a billion parameters while models like GPT-3 \citep{brown2020language}, for example, contain 175 billion parameters.

Finally, we only consider fake news as a binary classification task, though in reality the nature of news statements may be more complex. For example, a statement may be partially true and partially false, or it may be technically true but misleading.

\section*{Ethics Statement}

Our study is intended to highlight the importance of evaluating how well machine learning-based fake news detectors generalize to out-of-sample data. As described in the Limitations section, there are, of course, many existing datasets and models not included in this study. Thus, our result should not be used to automatically discredit those who use machine learning-based fake news detection methods in practice, for their robustness must be studied individually.

Furthermore, we acknowledge that the term, "Fake News," is often used in a dishonest way to criticize opponenets in politics. We discuss the definition of "Fake News" at the very beginning of our Introduction section to clarify that it is being used in a politically-neutral fashion to denote news that is false and may be created to intentionally spread lies.

\section*{Acknowledgements}

I thank Jason Lucas and Professor Dongwon Lee from Pennsylvania State University as well as Christian Classen from the University of Illinois Urbana-Champaign for their guidance and feedback.

\bibliography{anthology,custom}
\bibliographystyle{acl_natbib}

\appendix

\section{Out-of-Sample Matrices}

Below are the matrices for the out-of-sample performance of all deep and traditional models not included in the main text. The F1 scores are used in each cell.

\begin{figure}
    \includegraphics[width=.40\textwidth]{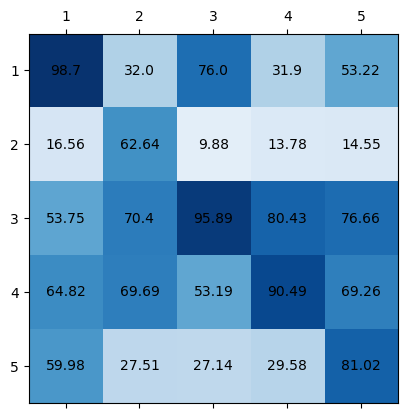}\hfill
    \caption{BERT}\label{fig:foobar}
\end{figure}

\begin{figure}
    \includegraphics[width=.40\textwidth]{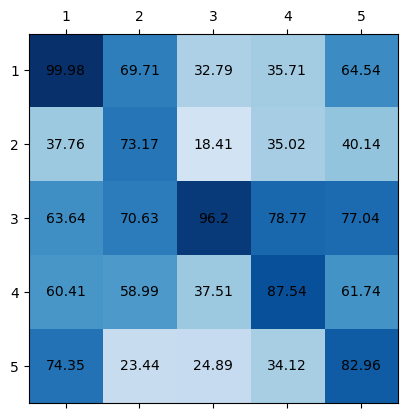}\hfill
    \caption{RoBERTa}\label{fig:foobar}
\end{figure}

\begin{figure}
    \includegraphics[width=.40\textwidth]{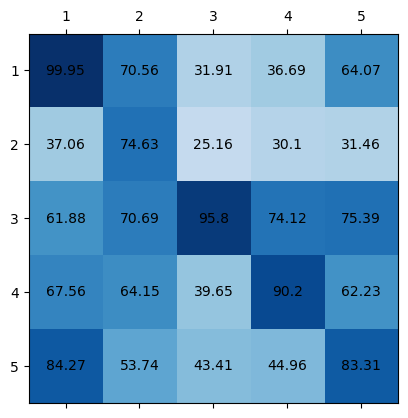}\hfill
    \caption{DeBERTa}\label{fig:foobar}
\end{figure}

\begin{figure}
    \includegraphics[width=.40\textwidth]{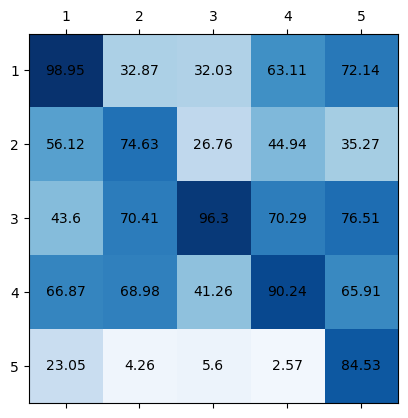}\hfill
    \caption{CT-BERT}\label{fig:foobar}
\end{figure}

\begin{figure}
    \includegraphics[width=.40\textwidth]{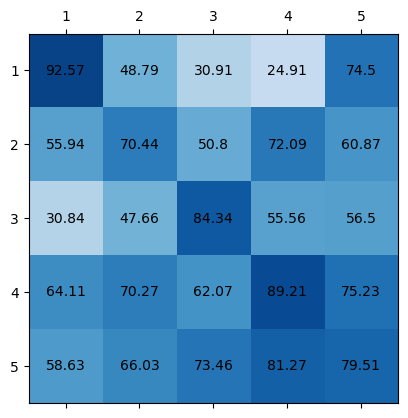}\hfill
    \caption{Naive Bayes}\label{fig:foobar}
\end{figure}

\begin{figure}
    \includegraphics[width=.40\textwidth]{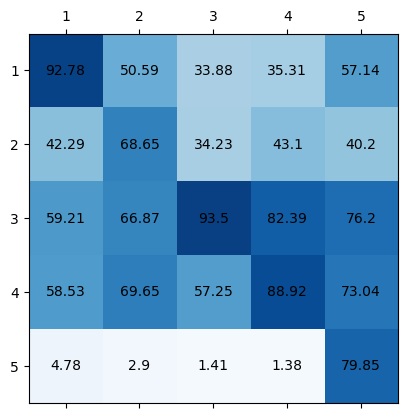}\hfill
    \caption{Random Forest}\label{fig:foobar}
\end{figure}

\begin{figure}
    \includegraphics[width=.40\textwidth]{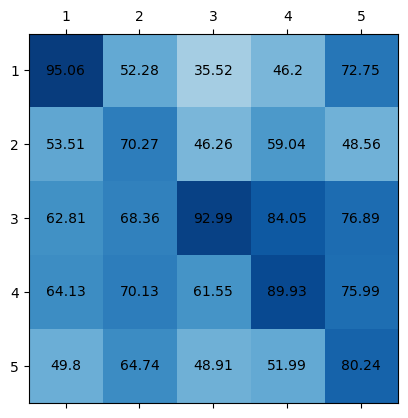}\hfill
    \caption{Support Vector Machine}\label{fig:foobar}
\end{figure}

\begin{figure}
    \includegraphics[width=.40\textwidth]{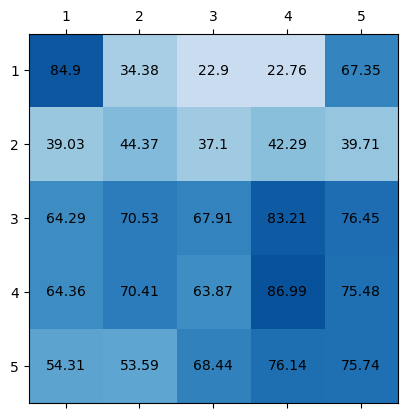}\hfill
    \caption{k-NN}\label{fig:foobar}
\end{figure}

\newpage

\section{Words Used in AdaBoost Classifiers}

The word used in each decision tree of the AdaBoost classifiers are shown below. The trees use the statistical importance of the words defined by the TF-IDF embeddings. Their training error is also included in each row.

\begin{table}
  \centering
  \caption{Words used in the AdaBoost classifier trained on D1 and the error of the corresponding decision tree}
  \begin{tabular}{|l|l|l|}
    \hline
     & Word & Classifier Error \\
    \hline
    1 & video & 0.3396 \\
    \hline
    2 & say & 0.4373 \\
    \hline
    3 & watch & 0.4596 \\
    \hline
    4 & hillary & 0.4587 \\
    \hline
    5 & obama & 0.4583 \\
    \hline
    6 & tweet & 0.4736 \\
    \hline
    7 & breaking & 0.4799 \\
    \hline
    8 & gop & 0.4811 \\
    \hline
    9 & trump & 0.4385 \\
    \hline
    10 & trump & 0.4471 \\
    \hline
    11 & house & 0.4761 \\
    \hline
    12 & north & 0.4819 \\
    \hline
    13 & america & 0.4845 \\
    \hline
    14 & donald & 0.4837 \\
    \hline
    15 & black & 0.4851 \\
    \hline
    16 & fatbox & 0.4872 \\
    \hline
    17 & news & 0.4845 \\
    \hline
    18 & image & 0.4889 \\
    \hline
    19 & get & 0.4844 \\
    \hline
    20 & room & 0.4887 \\
    \hline
    21 & china & 0.4867 \\
    \hline
    22 & eu & 0.4886 \\
    \hline
    23 & senate & 0.4857 \\
    \hline
    24 & minister & 0.4895 \\
    \hline
    25 & video & 0.4856 \\
    \hline
    26 & muslim & 0.4853 \\
    \hline
    27 & american & 0.4900 \\
    \hline
    28 & detail & 0.4866 \\
    \hline
    29 & seek & 0.4907 \\
    \hline
    30 & talk & 0.4895 \\
    \hline
    31 & court & 0.4875 \\
    \hline
    32 & liberal & 0.4906 \\
    \hline
    33 & urge & 0.4914 \\
    \hline
    34 & racist & 0.4916 \\
    \hline
    35 & pm & 0.4911 \\
    \hline
    36 & source & 0.4912 \\
    \hline
    37 & clinton & 0.4869 \\
    \hline
    38 & south & 0.4918 \\
    \hline
    39 & russia & 0.4877 \\
    \hline
    40 & tax & 0.4893 \\
    \hline
    41 & myanmar & 0.4932 \\
    \hline
    42 & wow & 0.4933 \\
    \hline
    43 & one & 0.4899 \\
    \hline
    44 & go & 0.4905 \\
    \hline
    45 & uk & 0.4931 \\
    \hline
    46 & cop & 0.4933 \\
    \hline
    47 & trump & 0.4621 \\
    \hline
    48 & official & 0.4900 \\
    \hline
    49 & ex & 0.4933 \\
    \hline
    50 & islamic & 0.4933 \\
    \hline
        
  \end{tabular}
\end{table}

\begin{table}
  \centering
  \caption{Words used in the AdaBoost classifier trained on D2 and the error of the corresponding decision tree}
  \begin{tabular}{|l|l|l|}
    \hline
     & Word & Classifier Error \\
    \hline
    1 & percent & 0.4379 \\
    \hline
    2 & obama & 0.4776 \\
    \hline
    3 & since & 0.4868 \\
    \hline
    4 & obamas & 0.4916 \\
    \hline
    5 & georgia & 0.4917 \\
    \hline
    6 & walker & 0.4919 \\
    \hline
    7 & rep & 0.4926 \\
    \hline
    8 & highest & 0.4933 \\
    \hline
    9 & average & 0.4930 \\
    \hline
    10 & obamacare & 0.4929 \\
    \hline
    11 & half & 0.4921 \\
    \hline
    12 & still & 0.4953 \\
    \hline
    13 & day & 0.4918 \\
    \hline
    14 & year & 0.4836 \\
    \hline
    15 & medicare & 0.4942 \\
    \hline
    16 & government & 0.4905 \\
    \hline
    17 & romney & 0.4945 \\
    \hline
    18 & mccain & 0.4945 \\
    \hline
    19 & million & 0.4943 \\
    \hline
    20 & nothing & 0.4971 \\
    \hline
    21 & face & 0.4977 \\
    \hline
    22 & debt & 0.4938 \\
    \hline
    23 & country & 0.4907 \\
    \hline
    24 & even & 0.4966 \\
    \hline
    25 & le & 0.4956 \\
    \hline
    26 & nearly & 0.4953 \\
    \hline
    27 & actually & 0.4947 \\
    \hline
    28 & plan & 0.4951 \\
    \hline
    29 & wisconsin & 0.4920 \\
    \hline
    30 & group & 0.4974 \\
    \hline
    31 & debate & 0.4972 \\
    \hline
    32 & whether & 0.4971 \\
    \hline
    33 & muslim & 0.4965 \\
    \hline
    34 & percent & 0.4958 \\
    \hline
    35 & well & 0.4972 \\
    \hline
    36 & time & 0.4965 \\
    \hline
    37 & spending & 0.4946 \\
    \hline
    38 & third & 0.4977 \\
    \hline
    39 & month & 0.4957 \\
    \hline
    40 & american & 0.4941 \\
    \hline
    41 & check & 0.4970 \\
    \hline
    42 & rank & 0.4976 \\
    \hline
    43 & stimulus & 0.4959 \\
    \hline
    44 & became & 0.4977 \\
    \hline
    45 & least & 0.4974 \\
    \hline
    46 & minute & 0.4980 \\
    \hline
    47 & clinton & 0.4944 \\
    \hline
    48 & line & 0.4979 \\
    \hline
    49 & almost & 0.4958 \\
    \hline
    50 & white & 0.4968 \\
    \hline
    
  \end{tabular}
\end{table}

\begin{table}
  \centering
  \caption{Words used in the AdaBoost classifier trained on D3 and the error of the corresponding decision tree}
  \begin{tabular}{|l|l|l|}
    \hline
     & Word & Classifier Error \\
    \hline
    1 & york & 0.2032 \\
    \hline
    2 & breitbart & 0.2349 \\
    \hline
    3 & hillary & 0.4436 \\
    \hline
    4 & time & 0.4575 \\
    \hline
    5 & border & 0.4789 \\
    \hline
    6 & trump & 0.4470 \\
    \hline
    7 & comment & 0.4857 \\
    \hline
    8 & election & 0.4807 \\
    \hline
    9 & migrant & 0.4853 \\
    \hline
    10 & islamic & 0.4882 \\
    \hline
    11 & texas & 0.4875 \\
    \hline
    12 & war & 0.4867 \\
    \hline
    13 & video & 0.4865 \\
    \hline
    14 & illegal & 0.4902 \\
    \hline
    15 & terror & 0.4907 \\
    \hline
    16 & american & 0.4915 \\
    \hline
    17 & pope & 0.4923 \\
    \hline
    18 & breaking & 0.4922 \\
    \hline
    19 & delingpole & 0.4934 \\
    \hline
    20 & house & 0.4891 \\
    \hline
    21 & gun & 0.4916 \\
    \hline
    22 & email & 0.4929 \\
    \hline
    23 & america & 0.4876 \\
    \hline
    24 & inaguration & 0.4934 \\
    \hline
    25 & cartel & 0.4933 \\
    \hline
    26 & breitbart & 0.4924 \\
    \hline
    27 & world & 0.4898 \\
    \hline
    28 & trump & 0.4888 \\
    \hline
    29 & anti & 0.4879 \\
    \hline
    30 & de & 0.4940 \\
    \hline
    31 & pen & 0.4940 \\
    \hline
    32 & latest & 0.4931 \\
    \hline
    33 & reason & 0.4945 \\
    \hline
    34 & york & 0.4808 \\
    \hline
    35 & time & 0.4854 \\
    \hline
    36 & sanctuary & 0.4944 \\
    \hline
    37 & york & 0.4857 \\
    \hline
    38 & housing & 0.4941 \\
    \hline
    39 & attack & 0.4911 \\
    \hline
    40 & wikileaks & 0.4943 \\
    \hline
    41 & fake & 0.4920 \\
    \hline
    42 & paris & 0.4936 \\
    \hline
    43 & parenthood & 0.4947 \\
    \hline
    44 & session & 0.4943 \\
    \hline
    45 & coulter & 0.4949 \\
    \hline
    46 & russia & 0.4895 \\
    \hline
    47 & report & 0.4879 \\
    \hline
    48 & profit & 0.4945 \\
    \hline
    49 & veteran & 0.4951 \\
    \hline
    50 & exclusive & 0.4945 \\
    \hline
        
  \end{tabular}
\end{table}

\end{document}